\documentclass[runningheads]{llncs}

\usepackage{accv}

\usepackage{accvabbrv}
\usepackage{graphicx,wrapfig}
\usepackage{amsmath}
\usepackage{epstopdf}
\usepackage{epsfig}
\usepackage{tikz,pgfplots}
\usepackage{amsmath}
\usepackage{stackengine}
\usepackage{physics}
\usepackage[outline]{contour} 
\usetikzlibrary{angles,quotes} 
\contourlength{1.2pt}

\tikzset{>=latex} 
\usepackage{xcolor}
\colorlet{veccol}{green!70!black}
\colorlet{vcol}{green!70!black}
\colorlet{xcol}{blue!85!black}
\colorlet{projcol}{xcol!60}
\colorlet{unitcol}{xcol!60!black!85}
\colorlet{myblue}{blue!70!black}
\colorlet{myred}{red!90!black}
\colorlet{mypurple}{blue!50!red!80!black!80}
\tikzstyle{vector}=[->,very thick,xcol]

\usepackage[pagebackref,breaklinks,colorlinks,citecolor=accvblue]{hyperref}

\usepackage{orcidlink}

\begin{document}

\title{CCNDF: Curvature Constrained Neural Distance Fields from  3D LiDAR Sequences} 

\titlerunning{Curvature Constrained Neural Distance Fields}

\author{Akshit Singh \and
Karan Bhakuni \and
Rajendra Nagar}

\authorrunning{A. Singh et al.}

\institute{ Indian Institute of Technology, Jodhpur, India
\email{\{singh.190,m22rm003,rn\}@iitj.ac.in}}

\maketitle

\begin{abstract}

Neural distance fields (NDF) have emerged as a powerful tool for addressing challenges in 3D computer vision and graphics downstream problems. While significant progress has been made to learn NDF from various kind of sensor data, a crucial aspect that demands attention is the supervision of neural fields during training as the ground-truth NDFs are not available for large-scale outdoor scenes. Previous works have utilized various forms of expected signed distance to guide model learning. Yet, these approaches often need to pay more attention to critical considerations of surface geometry and are limited to small-scale implementations. To this end, we propose a novel methodology leveraging second-order derivatives of the signed distance field for improved neural field learning. Our approach addresses limitations by accurately estimating signed distance, offering a more comprehensive understanding of underlying geometry. To assess the efficacy of our methodology, we conducted comparative evaluations against prevalent methods for mapping and localization tasks, which are primary application areas of NDF. Our results demonstrate the superiority of the proposed approach, highlighting its potential for advancing the capabilities of neural distance fields in computer vision and graphics applications.
\end{abstract}

\section{Introduction}
Neural distance fields (NDF) have gained significant attention for their remarkable contributions to the fields of 3D computer vision and robotics. The appeal of NDF lies in its ability to provide a continuous representation \cite{sitzmann2020implicit}, unhindered by grid resolution constraints \cite{chibane2020neural}, making it particularly versatile. Moreover, NDF accurately captures low-dimensional information and serves as a memory-efficient approach for mapping and utilization in mobile tasks \cite{locndf},\cite{shine}. Traditionally, NDF training involved either expensive calculations of ground truth signed distance values or relied on assumed ground truth information \cite{isdf}. While supervising NDF with ground truth meshes and signed distances laid crucial foundations for this line of study, obtaining such ground truth data for real-world tasks proves impractical.\\
To address this challenge and advance the applicability of NDF, recent developments have emerged. These novel methods focus on predicting signed distances with only point clouds as input, eliminating the need for ground truth supervision. Instead, they leverage various properties of NDF to approximate expected ground truth values \cite{zhang2021learning,chibane2020neural}. However, some of these methods introduce geometric assumptions and train with unrealistic ground truth values, potentially leading to problematic geometric generation from the models \cite{locndf,chibane2020neural,isdf}. \\
In the pursuit of overcoming inherent geometric limitations and ensuring a precise alignment of NDF properties with the estimated signed distance, we introduce a novel NDF supervision method based on the second-order derivative property of NDF. While higher-order derivatives have been increasingly explored for diverse applications, such as shape smoothing \cite{yang2021geometry}, the second-order properties of NDF have remained untapped in the context of supervision. The effectiveness of our approach is evaluated on two fundamental problems, mapping and localization, both of which extensively utilize NDF \cite{shine,locndf}. Our findings indicate an improvement over the current state-of-the-art geometrical techniques, affirming the superiority of our proposed method. The subsequent sections delve into a comprehensive discussion of the limitations of current methodologies, elucidate the methodology, and expound on the ideation behind our proposed approach.
\\
In sum we make the following contributions:
\begin{itemize}
    \item We present a novel neural distance field supervision approach by exploiting the second-order derivative property of neural distance field, marking an advancement in the exploration of higher-order derivatives.
    \item We develop a geometrical approach to ensure a precise alignment of neural distance field properties with the estimated signed distances which enhance the accuracy and reliability of neural distance field models.
    \item We demonstrate that the proposed approach attains state-of-the-art (SOTA) performance for mapping and localization on benchmark challenging datasets.
\end{itemize}
The rest of the paper is organized as follows. In Section \ref{sec:back}, we review the relevant literature and discuss the major research gaps. In Section \ref{sec:met}, we describe the proposed approach to solve the considered problem and discuss its geometric significance. In Section \ref{sec:res}, we present our results and compare the performance of the proposed approach with that of the state-of-the-art approaches.


\section{Related Work}\label{sec:back}
The use of Signed Distance Fields (SDF) has been prevalent in computer graphics and computational geometry for various applications for a significant period of time \cite{oleynikova2016signed,Park_2019_CVPR}. More recently neural fields have emerged as a prominent representation for modelling three-dimensional environments and objects, owing to seminal work done in this field \cite{mescheder2019occupancy,mildenhall2021nerf,oleynikova2017voxblox,zhang2021learning}. Neural Distance Field (NDF) extends the concept of SDFs by using neural networks to represent and learn the implicit function describing the shape of neural fields. Instead of using traditional mathematical formulations for surfaces, NDFs employ neural networks to model the signed distance function \cite{chibane2020neural}. \\
NDFs have demonstrated significant success in efficiently representing room-scale 3D environments and are particularly useful in real-time applications \cite{isdf,imap,nice,locndf}. The majority of the approaches that have employed NDF have utilised multi-layer perceptron (MLP) and supervised them either through density fields \cite{vizzo2021poisson}, normals \cite{sitzmann2020implicit,williams2021neural,zhang2021learning}, or directly through the distance field \cite{chibane2020neural}. In recent times, few approaches have begun to investigate the possibility of relating NDF directly to sensor readings. However, these approaches sometimes make harsh geometric assumptions \cite{locndf,shine,azinovic2022neural} or supervise NDF directly with sensor readings of small batches while ignoring the entire geometry leading to a distorted view \cite{isdf}, especially for large-scale environment representation.\\
Among the limited number of approaches that have directly utilised sensor readings from LiDAR for mapping and localization \cite{loam,legoloam,fastlio,fastlio2,kiss-icp,locndf}, only a few methods (SHINE-Mapping \cite{shine} and  LocNDF \cite{locndf}) have used implicit surface representation for large-scale environments. The remaining methods have concentrated on room-scale environments and used RGB-D data \cite{isdf,yan2021continual,imap,vasilopoulos2023hio,wang2023co}. \\
Other works have utilised multiple MLPs to map the environment \cite{reiser2021kilonerf},\cite{shine}. NDF has also been applied in the fields of motion planning \cite{chaplot2021differentiable}, localization \cite{locndf}, navigating in neural field \cite{adamkiewicz2022vision}, and learning neural fields for deformable objects \cite{yan2021continual}. \begin{figure}{}
	\centering
	\begin{tikzpicture}
		\draw[red!60!black] (4,-0.5) -- (6.37,0);
		\draw[green!60!black] (4,-0.5) -- (5.55,0.8);
		\draw[orange!60!black] (4,-0.5) -- (5.57,1.35);
		\draw[cyan!60!black] (4,-0.5) -- (4.66,0.88);
		\draw[magenta!60!black] (4,-0.5) -- (6.33,0.82);
		\draw[densely dashed,black!70!white] (5.9,1.5) -- (6.99,-0.25);
		\draw[black!70!white] (7,-0.3) -- (4,1.2);
		\draw[densely dotted,blue!70!black,-stealth] (4,-0.5) -- (6.97,-0.249) node[below] {Ray};
		\draw [black!70!black, xshift=4cm] plot [smooth, tension=1] coordinates { (2,2) (1.7,0.5) (4,-0.5)} node[below right] {Surface};
		\filldraw (6.37,0) circle (1pt);
		\filldraw (6.99,-0.25) circle (1pt);
		\filldraw (4,-0.5) circle (1pt);
		\draw[cyan!60!black,thick] (6.96,1.9) -- (7.2,1.9) node[right] {Distance along surface normal};
		\draw[green!60!black,thick] (6.96,1.5) -- (7.2,1.5) node[right] {True signed distance};
		\draw[orange!60!black,thick] (6.96,1.1) -- (7.2,1.1) node[right] {Proposed distance using the curvature of NDF};
		\draw[magenta!60!black,thick] (6.96,0.7) -- (7.2,0.7) node[right] {Distance along the normal in the closest surface direction};
		\draw[red!60!black,thick] (6.96,0.3) -- (7.2,0.3) node[right] {Distance to the nearest point in a batch};
		\draw[blue!60!black,thick] (6.96,-0.1) -- (7.2,-0.1) node[right] {Ray distance};
	\end{tikzpicture}
	\caption{Different methods for computing the signed distance to supervise NDF for a point on a ray beam.}
	\label{fig:1}
\end{figure}
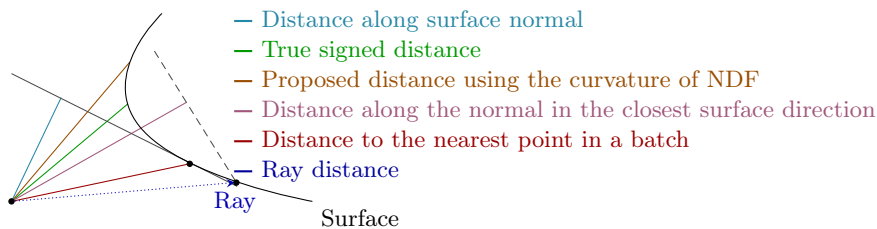
The objective of our research is to introduce a novel technique for effectively learning the NDF for large-scale environments by directly utilising the sensor readings, which can subsequently be applied to a range of tasks including path planning, mapping and localization, and more. This method leverages the geometric characteristics of the environment and also contributes to the current properties of NDF. In the following subsection, we have thoroughly examined the drawbacks of the existing methods.\\
\textbf{Geometric Drawbacks of Present Methods}: Several methodologies have been devised to supervise NDF with estimated signed distances. Earlier approaches often employed the ray distance as the approximated signed distance \cite{shine,azinovic2022neural}. Subsequent advancements in this line of research involved refining the supervision of NDF by incorporating the ray distance computed along the surface normal as the estimated signed distance \cite{isdf,zhang2021learning}. Recently, first-order derivative-based methodologies have been devised with the objective of calculating the ray distance along the normal pointing towards the closest surface \cite{locndf}. However, no method has yet investigated the higher-order properties of NDF to determine the expected NDF values and utilise them for network supervision.
In Fig. \ref{fig:1}, we tried to represent all the various distances that have been used for supervising MLP as estimated ground truth SDF values based on various methodologies. Fig. \ref{fig:1} illustrates the geometric limitations of different distances considered as ground truth signed distance values. For instance, using the nearest point to the sensor in a specific batch results in an underestimation of the ground truth signed distance values because it does not account for the geometry of the extended surface. Considering the distance along the surface normals as the true signed distance results in significant errors on a wide scale, as shown in Fig. \ref{fig:1} and also acknowledged by the authors of i-SDF \cite{isdf}. Considering the distance along the ray directly provides a simple option but is not particularly helpful for monitoring NDF as the ray distance is clearly not the minimal distance to the surface. i-SDF \cite{isdf} specifically designed to map room-scale environments directly calculated the closest distance to the surface in the selected batch and determined it as the global minimum distance to the surface.  Though this batch-wise technique was effective in a room-scale setting, it is deemed to be ineffective in a large-scale environment due to the extended geometry's impact on the NDF, which may not have been documented in the chosen batch.  Another seminal innovation in the estimation of the ground truth value was introduced by the authors of LocNDF \cite{locndf}, who explored the first-order derivative of NDF to determine the gradient to the nearest surface. They then calculated the ray distance along this determined gradient, resulting in a more accurate estimation of the signed distance value for linear surfaces. However, this method frequently produced inaccurate estimations for curvilinear or complex surfaces. As in Fig. \ref{fig:1}, we can clearly see that the distance along the normal pointing towards the closest surface overshoots the true signed distance field, resulting in error.\\
All other methods utilising NDF relied on high-resolution images or utilised ground truth SDF for NDF training \cite{chibane2020neural,vizzo2021poisson,sitzmann2020implicit}. These methods are not efficient for various purposes, particularly real-world applications, due to the lack of ground truth SDFs or high-resolution images. Instead, we only have sensor readings available to build the NDF. So for this work, we would directly use the raw LiDAR data for constructing NDF.\\
In this work, we introduced a new methodology for calculating the predicted SDF values by utilising higher-order derivatives of the NDF. We then compare this method with existing approaches used to estimate SDF values in large-scale environments. We specifically compare our results with ray distance and distance along the normal pointing towards the closest surface as these are the most widely used and latest approaches for determining signed distance values, moreover they also provide the tightest bound for approximating signed distances. Batch distance and normal distance along the surface normal are considered unsuitable for large-scale environments as they were originally designed for small-scale environments and are not theoretically applicable to large-scale settings. 
\section{Methodology}\label{sec:met}
This work focuses on learning the NDF from the LiDAR sensor readings in a self-supervised approach. Our approach eliminates the need for substantial processing of the sensor readings or creating specialised density fields or finding ground truth signed distance value manually, making it more practical for real-world applications like mapping and localization. Our methodology does not rely on point cloud normals as normal estimation has  inherent dependence on type of sensor, leading to potential errors in the process. The only information gained by the above mentioned sensors is the distance from the sensor's origin to the surface. In the following text, we will discuss how we exploited the properties of NDF and the environment's geometry to learn the NDF from the sensor reading.\\
\textbf{Problem Formulation and Background}: In this work, we address the problem of reconstruction of the implicit representation of the 3D surface of a scene from a sequence of LiDAR scans of the scene. Inspired from \cite{locndf}, we term the implicit surface as neural distance field (NDF). The NDF $\mathsf{D}$ maps a point $\mathsf{x}\in\mathbb{R}^3$ to a scalar value $\mathsf{D}(\mathsf{x})\in\mathbb{R}$ that represents the signed distance of the point $\mathsf{x}$ from the surface of the underlying scene. We use \cite{locndf} as a backbone with a better approximation of the distance of a point from the surface that we discuss later in this section. We first discuss the approach of LocNDF \cite{locndf} to create a context. LocNDF represents the NDF $\mathsf{D}$ as multilayer perceptron (MLP)  that consumes a point and returns it signed distance from the surface, i.e., it predicts $\mathsf{D}(\mathsf{x})$. Therefore, this MLP learns a representation of the 3D geometry of the scene. Instead of passing a point directly as an input to the MLP, its positional encoding \cite{mildenhall2021nerf} is being passed for better learning the 3D surface geometry. This encoding converts the coordinates into high-dimensional vectors by utilising periodic activation functions helping to store high dimensional information of points. The positional encoding  $\mathsf{P}:\mathbb{R}^3\rightarrow \mathbb{R}^{6h+3}$ of a point is defined in Equation \eqref{eqpe}.  \begin{eqnarray}
  \mathsf{P}(\mathsf{x})&=&\begin{bmatrix}\mathsf{x}&\sin(\omega_1\mathsf{x})&\cos(\omega_1\mathsf{x})&\cdots&\sin(\omega_h\mathsf{x})&\cos(\omega_h\mathsf{x})\end{bmatrix}^\top.  
  \label{eqpe}
\end{eqnarray}
In order to keep the surface thin,  we have used truncated signed distance field (TSDF) \cite{newcombe2011kinectfusion,vizzo2022vdbfusion} representation of the NDF. The structure of the multi-layer perceptron is inspired from the SHINE \cite{shine} to learn the NDF from LiDAR scans where the training is done in an unsupervised manner as we do not have the ground-truth NDF available. The LiDAR scanners measure the distance between its origin and the surface point and this distance is referred to as the ray distance. To train the MLP, LocNDF \cite{locndf} proposed a self-supervised technique where a set of points $\{\mathsf{x}_1,\ldots,\mathsf{x}_{n_i}\}$ are sampled on the ray between the sensor origin $\mathsf{o}_i$ and the ray endpoint $\mathsf{e}_i$ (the point where the ray strikes the scene surface) as defined in Equation \eqref{sampp} 
\begin{eqnarray}
 \mathsf{x}_\ell=(1-t_\ell)\mathsf{o}_i+t_\ell\mathsf{e}_i&\text{ and }&
 t_\ell=\frac{1}{0.9}\left(1-10^{\frac{\ell}{n_i-1}-1}\right),\;\ell\in\{1,\ldots,n_i\}.
 \label{sampp}
\end{eqnarray}
Now, in order to train the MLP to learn the underlying surface, we require the ground-truth signed distance of each sample point $\mathsf{x}_\ell$ from the surface. Since both surface and the signed distance of each sample point is unknown, it is a chicken-and-egg problem to solve. This is a major challenge to address to solve this problem efficiently. The LocNDF framework uses the ray distance $d_{\ell}=\|\mathsf{e}_i-\mathsf{x}_{\ell}\|_2$ to approximate the signed-distance for each sample point from the surface that itself we have learn. The approximated signed distance of the $\mathsf{x}_\ell$ is defined as 
\begin{eqnarray}
 \hat{d}_{\ell}&=&\frac{\mathsf{n}_\ell^\top(\mathsf{e}_i-\mathsf{x}_{\ell})}{\|\mathsf{n}_\ell\|_2}, 
 \label{distlocndf}
\end{eqnarray}
 where $\mathsf{n}_\ell$ represents the approximation of the direction towards the closest surface and LocNDF defined it as $\mathsf{n}_\ell=-\nabla_{\mathsf{x}}\mathsf{D}(\mathsf{x}_\ell)$. 
\subsection{Curvature-Constrained Neural Distance Fields}
\begin{figure}
	\centering
	\stackunder{\begin{tikzpicture}
			\node[anchor=south west,inner sep=0] at (0,0) {\includegraphics[width=0.41\linewidth]{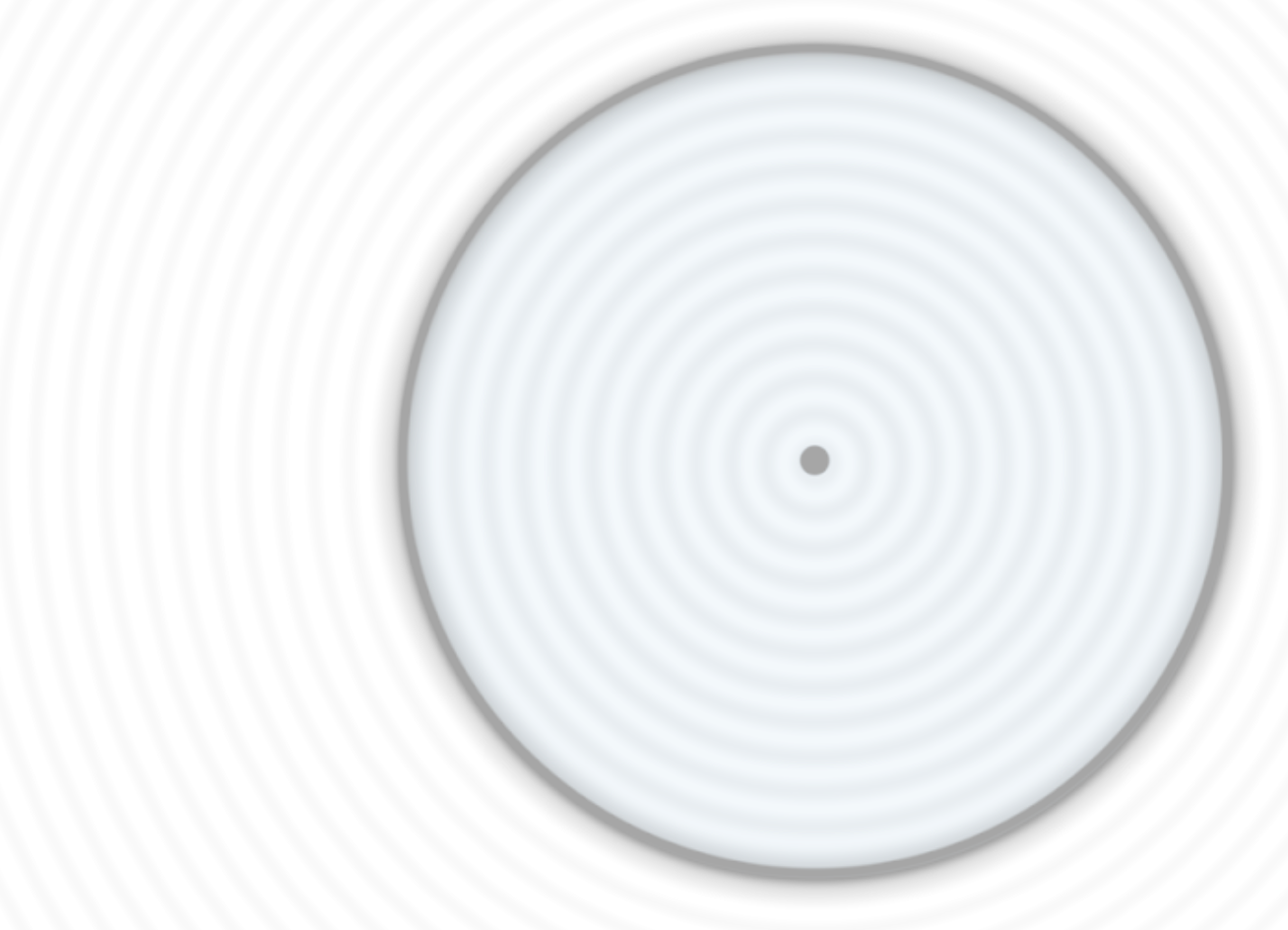}};
			\draw[cyan!90!black] (0.9,1.3) --	 (1.6,1.47);
			\draw[cyan!90!black] (0.1,1.53) -- (3.85,0.37);
			\draw[cyan!90!black] (1.85,1.0) -- (3.176,1.826);
			\draw[red!90!black,dash dot] (0.565,3.0) -- (1.15,0.2);
			\draw[red!90!black,dash dot] (2.87,3.3) -- (3.176,1.826);
			\draw[red!90!black, dash dot] (1.6,1.47) -- (1.81,0.4);
			\draw[dashed,<->] (4.00,0.5) -- (3.176,1.826) node[scale=0.9,midway,below,rotate=-55] {\scriptsize ROC $r_\ell$};
			\draw[dashed,<->] (0.63,2.6) -- (2.92,3.13) node[scale=0.9,midway,above,rotate=12] {\scriptsize NDF ROC $R_\ell$};
			\draw[dashed,<->] (1.15,0.45) -- (1.75,0.6) node[scale=0.9,midway,below,rotate=12] { $\hat{d}_\ell$};
			\draw[projcol,<-] (1.6,1.47) -- (3.176,1.826) node[scale=0.9,midway,above,rotate=12] {\scriptsize $\mathsf{n}_{\ell}=-\nabla_{\mathsf{x}}\mathsf{D}(\mathsf{x}_\ell)$};
			\filldraw[blue!30!black,thick] (3.176,1.826) circle (1pt) node[right] {$\mathsf{c}_\ell$};
			\filldraw[blue!30!black,thick] (0.9,1.3) circle (1pt) node[below] {$\mathsf{x}_\ell$};
			\filldraw[blue!30!black,thick] (1.6,1.47) circle (1pt) node[below] {F};
			\filldraw[blue!30!black,thick] (0.1,1.53) circle (1pt) node[below] {A};
			\filldraw[blue!30!black,thick] (1.85,1.0) circle (1pt) node[below] {$\mathsf{e}_i$};
	\end{tikzpicture}}{(a)}
	\stackunder{\begin{tikzpicture}
			\node[anchor=south west,inner sep=0] at (0,0) {\includegraphics[width=0.56\linewidth]{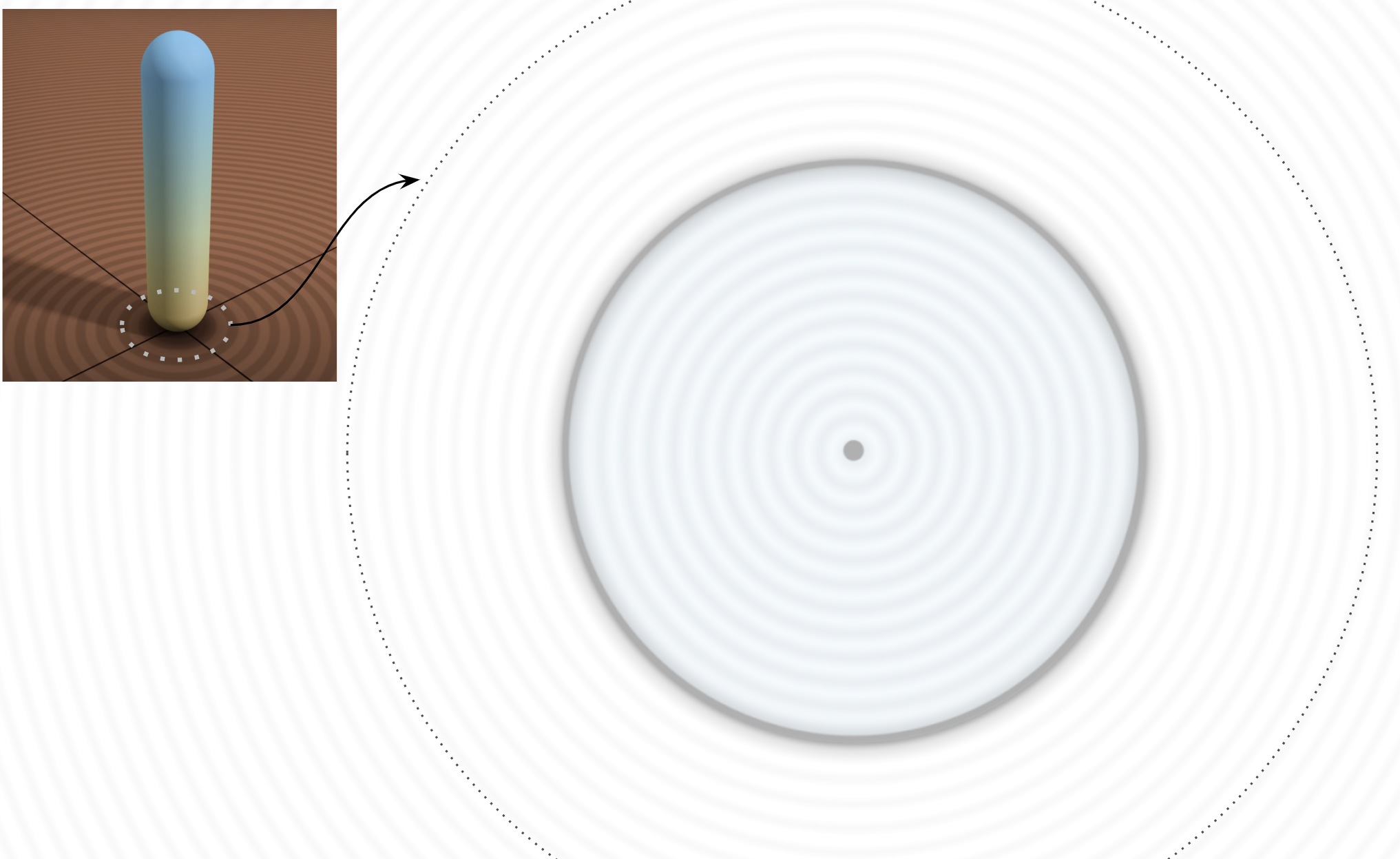}};
			\draw[cyan!90!black] (1.7,1.991) -- (3.2,1.991);
			\draw[cyan!90!black] (3.2,1.991) -- (4.17,1.991);
			\draw[cyan!90!black] (3.2,1.991) -- (3.2,1.0);
			\draw[cyan!90!black] (1.7,1.991) -- (3.2,1.0);
			\draw[red!90!black,dash dot] (2.75,1.991) -- (2.75,0.9);
			\draw[red!90!black,dash dot] (1.7,2.991) -- (1.7,0.9);
			\draw[red!90!black,dash dot] (3.2,1.991) -- (3.2,2.7);
			\draw[red!90!black,dash dot] (4.17,1.991) -- (4.17,3.5);
			\draw[red!90!black,dash dot] (1.7,1.991) -- (0.89,0.6);
			\draw[red!90!black,dash dot] (3.2,1.0) -- (2.68,0.08);
			\draw[dashed,<->] (1.7,2.2) -- (3.2,2.2) node[scale=0.9,midway,above,rotate=0] {\scriptsize DCN $=1.75$};
			\draw[dashed,<->] (1.7,1.13) -- (2.75,1.13) node[scale=0.9,midway,above,rotate=0] {\scriptsize TSD $=1$};
			\draw[dashed,<->] (1.7,2.7) -- (4.17,2.7) node[scale=0.9,midway,above,rotate=0] {\scriptsize NDF ROC $=2$};
			\draw[dashed,<->] (1.32,1.31) -- (2.78,0.3) node[scale=0.9,midway,below,rotate=-34] {\scriptsize Ray Dist. $=1.4$};
			\draw[dashed,<->] (4.17,1.991) -- (4.17,0.591) node[scale=0.9,midway, right,rotate=0] {\scriptsize ROC of Surface $=1$};
			\draw[dashed] (2.4,3.8) -- (2.4,3.8) node[scale=0.9, right,rotate=0] {\scriptsize $x^2+y^2=4$};
			\draw[unitcol,dashed] (4.3,3.5) -- (4.3,3.5) node[scale=0.9, right,rotate=0] {\scriptsize $x^2+y^2=1$};
			\filldraw[blue!30!black,thick] (1.7,1.991) circle (1pt) node[left] {$\mathsf{x}_\ell$};
			\filldraw[blue!30!black,thick] (4.17,1.991) circle (1pt) node[right] {$\mathsf{c}_\ell$};
			\filldraw[blue!30!black,thick] (3.2,1.991) circle (1pt) node[below right] {F};
			\filldraw[blue!30!black,thick] (2.75,1.991) circle (1pt);
			\filldraw[blue!30!black,thick] (3.2,1.0) circle (1pt) node[below right] {$\mathsf{e}_i$};;
	\end{tikzpicture}}{(b)}
\caption{The figure illustrates the concentric nature of NDF; ROC of points on NDF constantly increase as we move away from the surface}
\label{fig2:overall}
\end{figure}
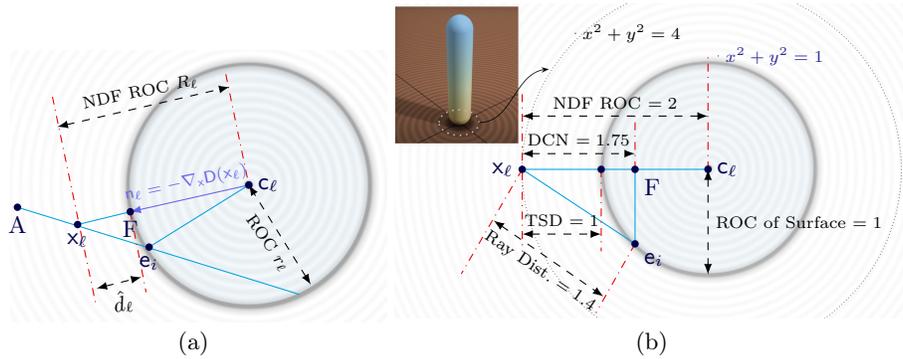

\textbf{Proposed Approach}: In order to estimate the distance to the nearest surface, we utilise the characteristics of NDF. Given that the NDF represents the distance of a point to the closest surface, and the iso-lines of the NDFs are concentric to the surface of object, indicating that point $\mathsf{x}_\ell$ will lie on isolines of the NDF. Isolines represent the NDF and are formed by the distance to the nearest surface. Fig. \ref{fig2:overall}(a) shows that the object's NDF are concentric, and the radius of curvature (ROC) of isolines increases in magnitude as one moves away from the surface. 
Based on this intuition, the  radius of curvature (ROC) of the isoline at a given point is  the ROC of that point on the NDF. As isolines are only formed to represent the distance to the nearest surface, this would ensure that the ROC of a point on the iso-line would be concentric with the ROC of the corresponding point that is the nearest point on the surface from the inquiry point. Consequently, if the ROC of the corresponding point on the surface could be determined, the difference between the two would provide a reliable approximation of the signed distance. While NDFs can be affected by multiple surfaces, discontinuities typically arise at significant distances. When closer to the surface, NDFs tend to be concentric to the inquiry point \cite{takikawa2022dataset}. To mitigate discontinuity effects, we have sampled points log-linearly along the LiDAR ray, as defined in Equation \eqref{sampp}, assigning higher weights to those closer to the surface. This approach efficiently reduces discontinuity impacts on distance calculation. Now, according to \cite{curvature}, the ROC for any point in the implicit representation is defined as $R_{\ell}=\frac{1}{|\kappa_\ell|}$. Here, $\kappa_\ell$ is the mean curvature or Germain curvature \cite{holmes2023germain} at a point $\mathsf{x}_\ell$ on the implicit surface and is defined as:
\begin{equation}
    \kappa_\ell= \frac{\nabla^\top_{\mathsf{x}}\mathsf{D}(\mathsf{x}_\ell) \mathsf{H}(\mathsf{x}_\ell)\nabla_{\mathsf{x}_\ell}\mathsf{D}(\mathsf{x}_\ell)-\|\nabla_{\mathsf{x}}\mathsf{D}(\mathsf{x}_\ell)\|_2^2 \text{trace}(\mathsf{H}(\mathsf{x}_\ell))}{2\|\nabla_{\mathsf{x}}\mathsf{D}(\mathsf{x}_\ell)\|_2^3}.\label{eqk}
\end{equation}
The mean curvature $\kappa_\ell$ can be found by Equation \eqref{eqk} for calculating the ROC for implicit surfaces if we have the analytical expression of $\mathsf{D}$ available with us. The NDF is theoretically described as a scalar field representing the extrinsic properties of the scene. The formula described in Equation \eqref{eqk} is intricate and computationally demanding. A more computationally efficient way \cite{curvature} to calculate the mean curvature is defined in Equation \eqref{eqc}. 
\begin{equation}
    \kappa_\ell= \left|\nabla^\top_{\mathsf{x}} \left(\frac{\nabla_{\mathsf{x}}\mathsf{D}(\mathsf{x}_\ell)}{\|\nabla_{\mathsf{x}_\ell}\mathsf{D}(\mathsf{x})\|_2}\right)\right|=\frac{1}{R_\ell}\Rightarrow R_\ell=\left|\nabla^\top_{\mathsf{x}} \left(\frac{\nabla_{\mathsf{x}}\mathsf{D}(\mathsf{x}_\ell)}{\|\nabla_{\mathsf{x}_\ell}\mathsf{D}(\mathsf{x})\|_2}\right)\right|^{-1}.
    \label{eqc}
\end{equation}
Now, we establish a relationship between the ROC of a point and its closet point on the surface. As depicted in Fig. \ref{fig2:overall} (a), we  observe  that the NDF comprises of concentric spheres and when we apply the Equation \eqref{eqc} at any sampled enquiry point $\mathsf{x}_\ell$ on the LiDAR bean having an origin at $A$, we receive the ROC of isoline at $\mathsf{x}_\ell$. As a property of NDF, we are assured that this ROC is actually concentric with ROC of the point on the surface nearest to the enquiry point.  Here the corresponding point on the surface to the enquiry point $\mathsf{x}_\ell$ is point $F$. The formation of the neural field at point $\mathsf{x}_\ell$ is solely due to point $F$ on the surface, as the NDF is a representation of the distance to the nearest point. \\
Let the differential ROC of isoline at point $\mathsf{x}_\ell$ be $R_\ell$, and let $\mathsf{e}_i-\mathsf{x}_\ell$ be the known ray distance to the surface, where $\mathsf{e}_i$ is the point on the surface at which LiDAR ray intersects. Now the approximated signed distance value is $\|\mathsf{x}_\ell-F\|_2$, i.e., the difference of ROC of the enquiry point and the corresponding point. Using Equation \eqref{eqc},  we know the radius $R_\ell$. We exploit the geometry present to calculate the ROC $r_\ell$ at the corresponding point $F$. In Fig. \ref{fig2:overall} (a), we apply the cosine rule in triangle $\displaystyle \bigtriangleup \mathsf{x}_\ell\mathsf{e}_i\mathsf{c}_\ell$ as follows:
\begin{eqnarray}
    \|\mathsf{e}_i-\mathsf{c}_\ell\|_2^2&=& \|\mathsf{e}_i-\mathsf{x}_\ell\|_2^2 + \|\mathsf{c}_\ell-\mathsf{x}_\ell\|_2^2-2\|\mathsf{e}_i-\mathsf{x}_\ell\|_2 \|\mathsf{c}_\ell-\mathsf{x}_\ell\|_2\cos\theta\\
    r^2_\ell&=&d_\ell^2+R_\ell^2-2d_\ell R_\ell\frac{\mathsf{n}^\top_\ell(\mathsf{e}_i-\mathsf{x}_\ell)}{\|\mathsf{e}_i-\mathsf{x}_\ell\|_2}=d_\ell^2+R_\ell^2-2R_\ell\mathsf{n}^\top_\ell(\mathsf{e}_i-\mathsf{x}_\ell).
    \label{9}
\end{eqnarray}
Here, $\theta=\angle{\mathsf{c}_\ell\mathsf{x}_\ell\mathsf{e}_i}$. Our proposed estimation $\hat{d}_\ell$ for the ground-truth signed distance of the point $\mathsf{x}_\ell$ from the surface is defined in Equation \eqref{estd}.
\begin{eqnarray} 
    \hat{d}_\ell=R_\ell-r_\ell
    \Rightarrow\hat{d}_\ell&=&R_\ell-\sqrt{d_\ell^2+R_\ell^2-2R_\ell\mathsf{n}^\top_\ell(\mathsf{e}_i-\mathsf{x}_\ell)}.
    \label{estd}
\end{eqnarray}
We use this $\hat{d}_\ell$ to supervise the training of a multi-layer perception (MLP) to learn the NDF. In Fig. \ref{fig2:overall}(b), we have presented a toy example. By observing the projected NDF, which forms a circle at the inspection plane. The ray distance (SHINE mapping) is 1.41 that we directly obtain from the LiDAR measurements, the distance along the normal pointing toward the closest surface (DCN) proposed by LocNDF \cite{locndf} is 1.75 (Equation \eqref{distlocndf}), ROC of NDF is equal to 2, approximated ROC of the surface is equal to 1.02 (Equation \eqref{eqc}), the distance using our method is equal to 0.98, and the true signed distance (TSD) is equal to 1. We observe that the proposed distance (0.98) is very close to the TSD (1) as compared to the Ray Distance (1.41) and the distance proposed by LocNDF.\\ Now, since the learning of NDF is supervised from estimated $\hat{d}_\ell$ and $\hat{d}_\ell$ is estimated from the learned NDF. This is a chicken-and-egg problem to solve. Therefore, better we estimate $\hat{d}_\ell$,  better we will learn NDF and vice-versa. Due to this, the training of the MLP may not be stable as we initialize the MLP with random weights leading to random SDF but neither we nor LocNDF \cite{locndf} witnessed this situation. The rationale for not observing any instability was that the estimated error 
\begin{eqnarray}
\epsilon_{\ell}=\mathsf{D}(\mathsf{P}(\mathsf{x}_\ell))-R_\ell-\sqrt{d_\ell^2+R_\ell^2-2R_\ell\mathsf{n}^\top_\ell(\mathsf{e}_i-\mathsf{x}_\ell)}
\end{eqnarray}
 approaches zero when the query point $\mathsf{x}_{\ell}$ is sampled closer to the surface, since $\mathsf{D}(\mathsf{P}(\mathsf{x}_\ell))$ converge to zero, leading $\hat{d}_\ell$ to converge to zero as well. To ensure this,  we give higher weight $w_\ell=(d_{max}-\mathsf{D}(\mathsf{P}(\mathsf{x}_\ell)))^\gamma$ to the points  $\mathsf{x}_\ell$ with lower $\mathsf{D}(\mathsf{P}(\mathsf{x}_\ell))$, as proposed in \cite{locndf,vizzo2022vdbfusion}. Here, $d_{max}$ is the largest distance in the batch, while $\gamma$ is a parameter that determines the degree of weighting given to the nearby point. A larger value results in a greater influence from the nearby point.\\
Now to supervise the MLP to learn the NDF, we define the loss function in Equation \eqref{losseq} where we use the proposed estimation of the signed distance $\hat{d}_\ell$ for ray points $\mathsf{x}_\ell$ along with the regularizers inspired from LocNDF \cite{locndf}.
\begin{eqnarray}\label{loss}
\mathcal{L}=\sum_\ell\frac{w_\ell\epsilon_\ell}{\sum_jw_j}+\lambda_{1}\sum_\ell|\mathsf{D}(\mathsf{e}_\ell)|+\lambda_{2}\sum_\ell|\|\mathsf{n}_\ell\|_2-1|+\lambda_{3}\sum_\ell\sum_{j\in\mathcal{N}_\ell}|\mathsf{n}^\top_\ell\mathsf{n}_j|.
\label{losseq}
\end{eqnarray}
Here,  $\mathsf{n}_{j}$ is the normal at the neighboring point $\mathsf{x}_j\in\mathcal{N}_\ell$ of  $\mathsf{x}_\ell$, and $\lambda_{1},\lambda_{2}$ and $\lambda_{3}$ are hyper-parameters that we empirically determine.
\subsection{Geometric Advantages of Proposed Approach}
Equation  \eqref{estd} holds significance even for complex and intricate surfaces as these equations are defined for the differential element of NDF at a point.  Previous methods for estimating signed distances, like ray distances and ray distance along the surface normal, only considered the characteristics of the endpoint on the surface ($\mathsf{e}_{i}$) to estimate the signed distance. Theoretically, the NDF should not be dependent on this point because it might not be the nearest point on the surface to the enquiry point; therefore, it is problematic to have an NDF that is dependent on the characteristics of this point ($\mathsf{e}_{i}$). The ray distance along the normal to the nearest surface incorporates certain attributes extracted from the nearest point on the surface to the enquiry point. By determining the direction to the closest point and then estimating the signed distance, this method supposedly improves NDF supervision by making the method  reliant on the nearest point on the surface. However, this approach solely depended on the directional (vector) characteristics of the closest point on the surface, resulting in the estimated signed distance being influenced by the directional attributes of the closest point on the surface only.  Our method estimated the signed distance as the difference between the ROC at the enquiry point and the point closest to the enquiry point on the surface.  As a virtue of our method, the NDF became more dependent on the scalar property of the nearest point on the surface to the enquiry point, which is directly accountable for generating the neural field at that particular location. The dependence of NDF on the scalar properties of the closest surface point to the enquiry point enhances the reliability and accuracy of NDF supervision. \\
For example, in the geometry shown in the inset figure, if the ray intersects at point $B$, methods that rely on calculating signed distances based on attributes of the endpoint ($B$), such as ray distance and ray distance along the surface normal,\begin{wrapfigure}{r}{0.30\textwidth}\centering
	\begin{tikzpicture}
		\node[anchor=south west,inner sep=0] at (0,0) {\includegraphics[width=0.67\linewidth]{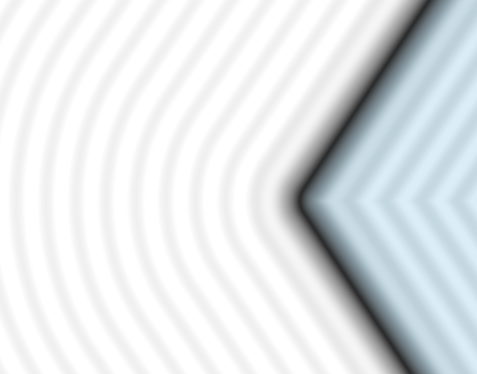}};
		\draw[orange!90!black] (0.5,0.88) -- (1.92,0.88);
		\draw[green!90!black] (1.15,1.5) -- (1.92,0.3);
		\draw[blue!90!black] (0.5,0.88) -- (1.15,1.5);
		\draw[red!90!black] (1.92,0.3) -- (0.5,0.88);
		\draw[magenta!70!black] (1.92,0.3) -- (1.92,0.88) ;
		\filldraw[blue!30!black,thick] (0.5,0.88) circle (1pt) node[left] {A};
		\filldraw[blue!30!black,thick] (1.92,0.88) circle (1pt) node[right] {D};
		\filldraw[blue!30!black,thick] (1.92,0.3) circle (1pt) node[right] {B};
		\filldraw[blue!30!black,thick] (1.55,0.88) circle (1pt) node[below] {C};
		\filldraw[blue!30!black,thick] (1.15,1.50) circle (1pt) node[right]  {E};
	\end{tikzpicture}
\end{wrapfigure} result in significant geometric errors. Conversely, using ray distance along the normal to the closest surface overestimates the distance by presuming the geometry to be linear, as it only examines  the direction to the  closest surface for estimating the distance.  Whereas, our method relies on the concentric properties  of NDF to calculate the concentric ROC at points $A$ and $C$, resulting in the most accurate estimate for the signed distance represented as segment $AC$. It is important to highlight that our method heavily depends on point $C$ and its magnitude properties, which are the actual cause of the neural field at point $A$, allowing us to incorporate extended geometric characteristics that would not be feasible if we only focused on properties of  $B$.
\section{Results and Evaluation}\label{sec:res}
Evaluating a geometric approach presents inherent challenges, particularly in the absence of ground truth data for metric verification. To address this complexity, our evaluation strategy involved comparing our approach with other relevant methodologies on two fundamental problems: mapping and localization. This allowed us to obtain insights from both qualitative and quantitative perspectives. Given that geometric techniques within NDF are employed for diverse purposes, from motion planning to localization, and each model utilizing these approaches serves distinct objectives, direct model-to-model comparisons may not be the most suitable approach. Instead, our evaluation strategy involved comparing different geometric approaches under unified conditions. This allowed for a more appropriate and insightful assessment, considering the diverse applications and objectives of these methodologies within the NDF framework.\\
\textbf{Training Setup.} For all the experiments conducted in this study, consistent parameters were employed. The positional encoding parameter, denoted as $h$, was set to 30. The implementation of the MLP utilized the SIREN architecture \cite{sitzmann2020implicit}, with the hidden feature dimension size set to 128. During training, $n_{i}$, representing the number of intervals between the sensor's starting point and endpoint, was set to 40. The coefficients for various loss components, as indicated loss equation \ref{loss}, were set as follows: $\alpha_{1}=10^{-1}$, $\alpha_{2}=10^{-4}$, $\alpha_{3}=10^{-3}$, and $\gamma=3$. The optimization algorithm utilized is AdamW, with an initial learning rate set to $10^{-4}$. The experiments have been developed on a desktop PC with Intel @ Xeon(R) Gold 6226R CPU @ 2.90GHz × 32 and an Nvidia RTX A6000.\\
\textbf{Mapping}: We conducted evaluation of our method on the Apollo Southbay ColumbiaPark-3 mapping run \cite{appolo} and the KITTI Sequence \cite{geiger2012we}, utilizing provided poses for different batches of scans, each comprising over 700 scans. Our approach involves direct training of the MLP for the entire scene, utilizing a bounding box of size 50m. The network was trained for 10 epochs, a process that took approximately 20 minutes. Scan registration followed the same procedure as employed by LocNDF \cite{locndf}. For visualization purposes, we solely utilized the mesh obtained from marching cubes \cite{marching}, and not for registration. The evaluation involved a comparative analysis of our results with other geometrical distances that have been previously employed in similar contexts. \begin{figure}[t]
    \centering
        \includegraphics[width=0.325\linewidth]{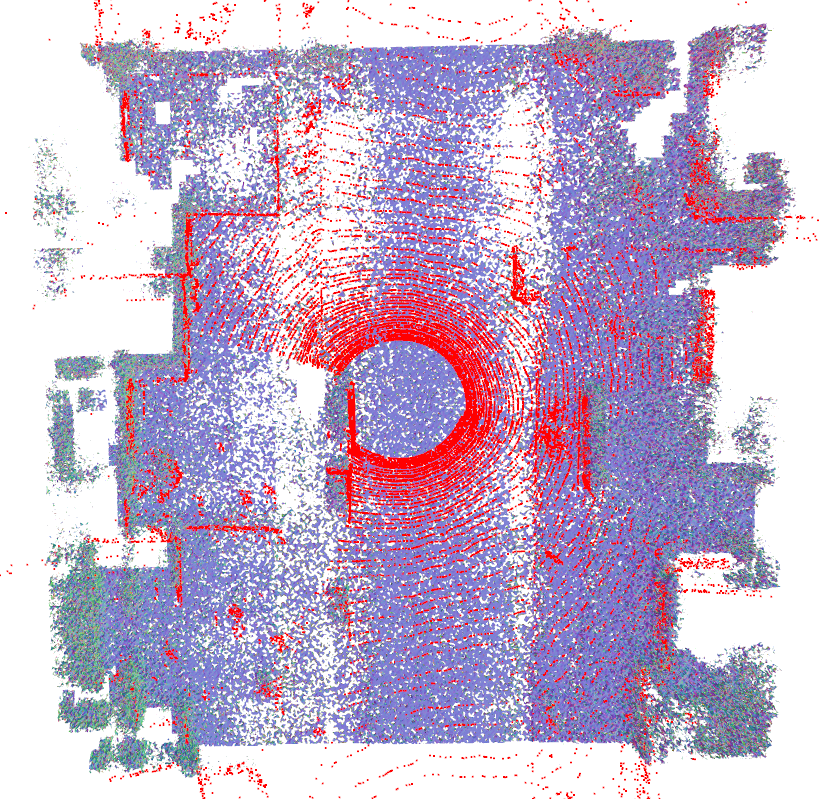}
        \includegraphics[width=0.325\linewidth]{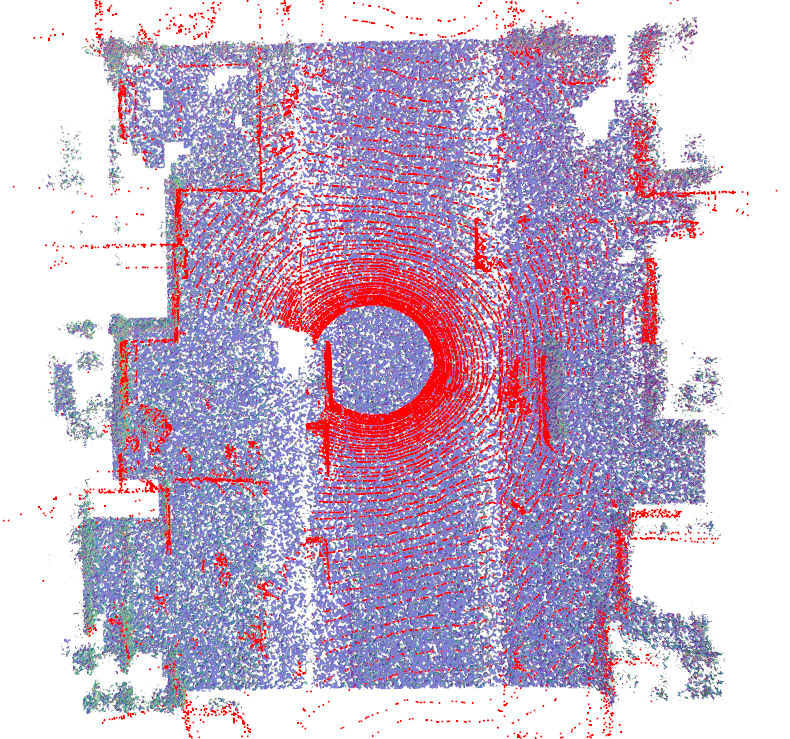}
        \includegraphics[width=0.325\linewidth]{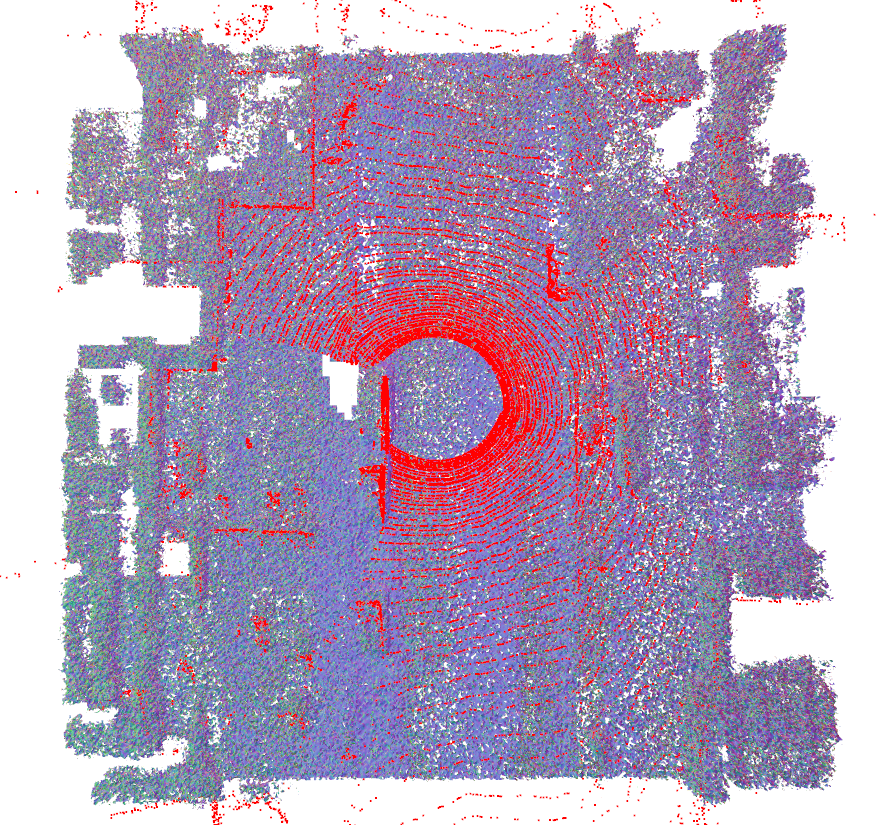}
        \includegraphics[width=0.325\linewidth]{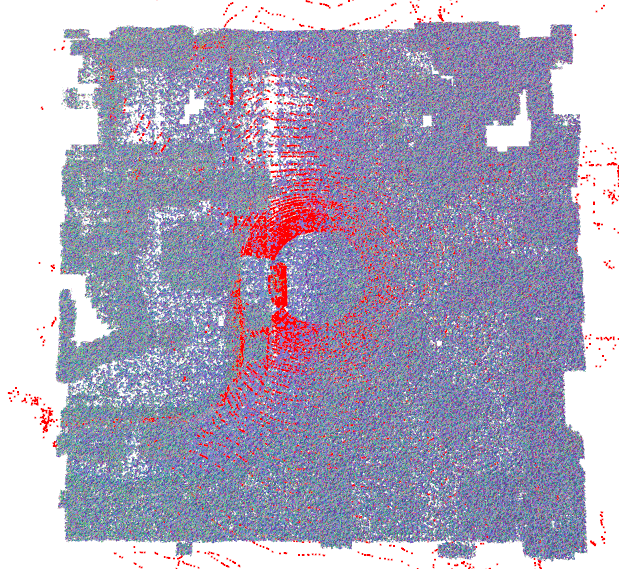}
        \includegraphics[width=0.325\linewidth]{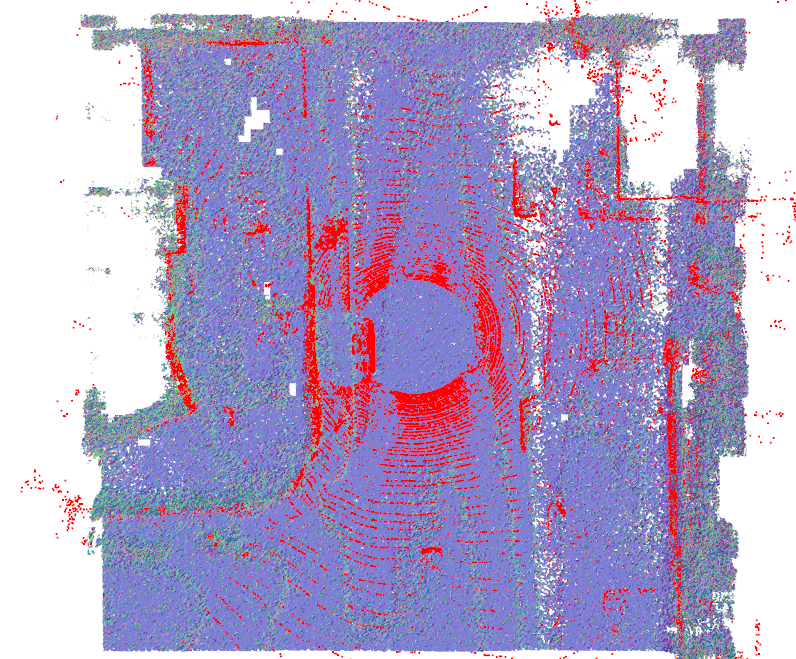}
        \includegraphics[width=0.325\linewidth]{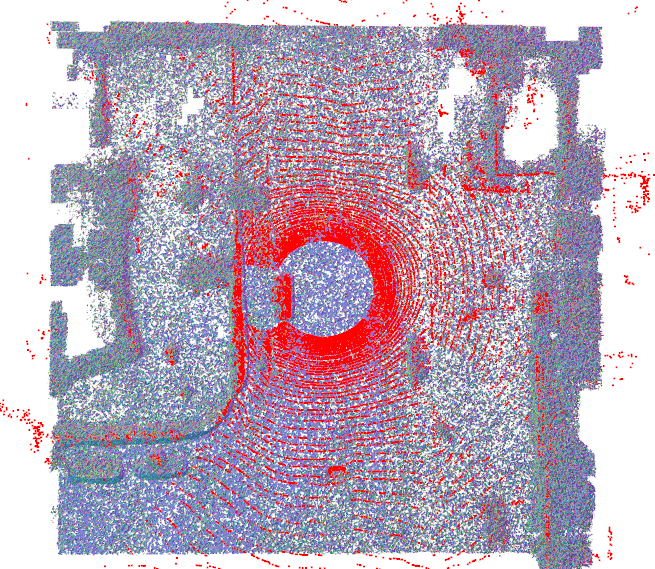}
        \stackunder{\includegraphics[width=0.325\linewidth]{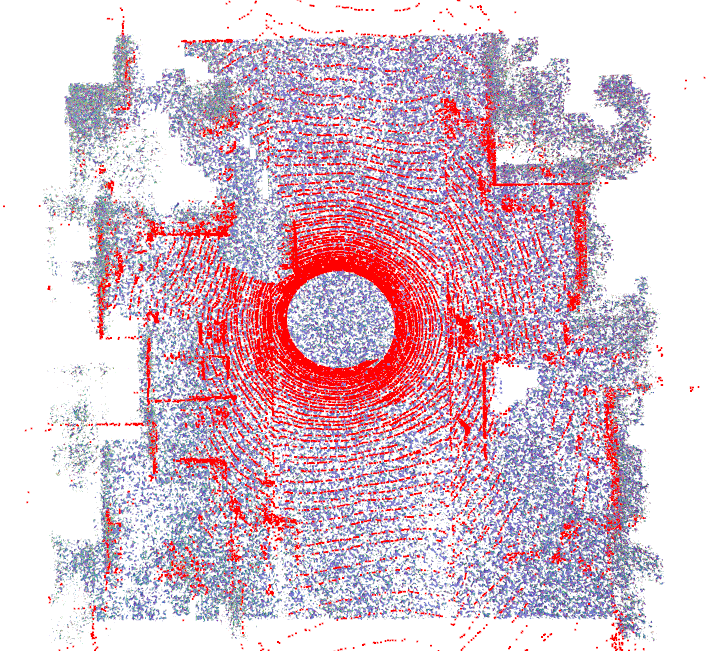}}{(a)}
        \stackunder{\includegraphics[width=0.325\linewidth]{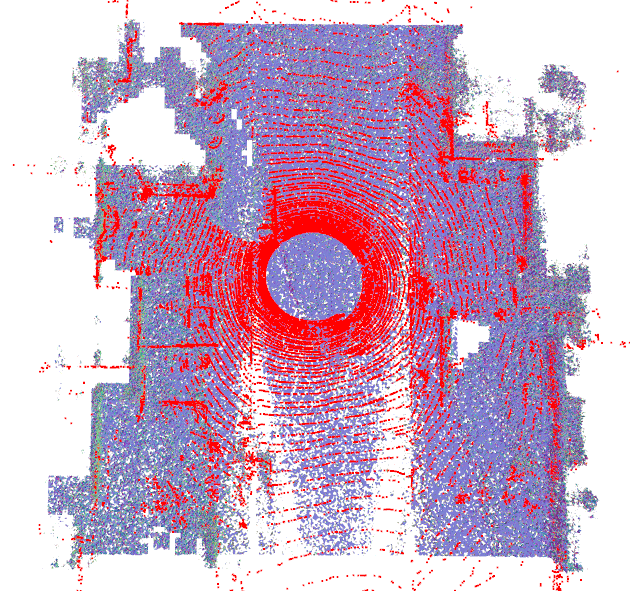}}{(b)}
        \stackunder{\includegraphics[width=0.325\linewidth]{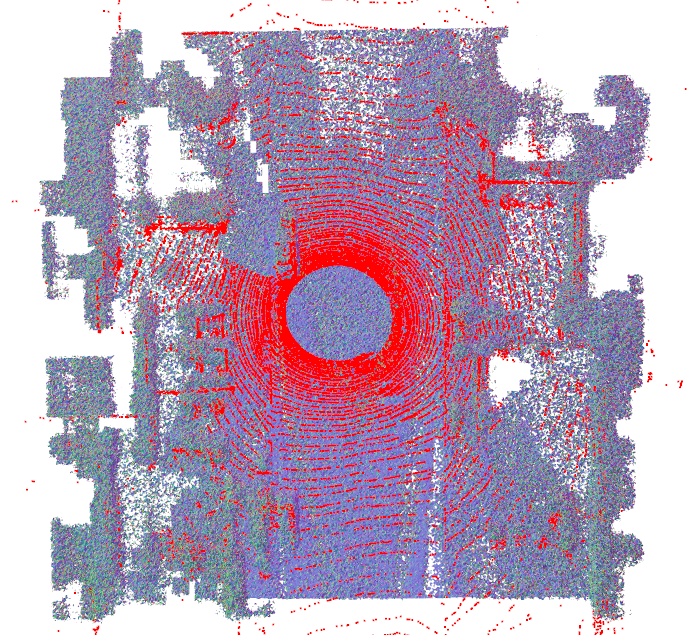}}{(c)}
    \caption{Visual comparison of results obtained with various geometrical methods for NDF supervision on the benchmark dataset. (a) Ray Distance, (b) Ray distance along the normal pointing towards the closest surface, and  (c) Proposed approach.}
    \label{fig:4}
\end{figure} In Fig. \ref{fig:4} and Fig. \ref{fkitti}, we present a qualitative comparison highlighting the geometric enhancements achieved by our method on different sequences from the Apollo ColumbiaPark-3 dataset \cite{appolo} and KITTI dataset \cite{geiger2012we}, respectively. The results clearly illustrate the limitations of supervision by ray distance, where the representation lacks detail and exhibits significant gaps. In the mapping obtained through supervision by ray distance along the normal towards the closest surface, we observe an improvement with finer details, a clearer path, and visible surrounding buildings. The results of our approach demonstrate a significant increase in detail. Our model effectively captures finer details, such as cars and nearby trees, that were previously missing. The visual comparison across all methods highlights the effectiveness of our proposed model. Fig. \ref{5} represents the front view mapping of the sequence showcasing geometrical features captured by different approaches on the Southbay ColumbiaPark-3 dataset \cite{appolo}.
\begin{figure}[t]
	\stackunder{\includegraphics[width=0.328\linewidth]{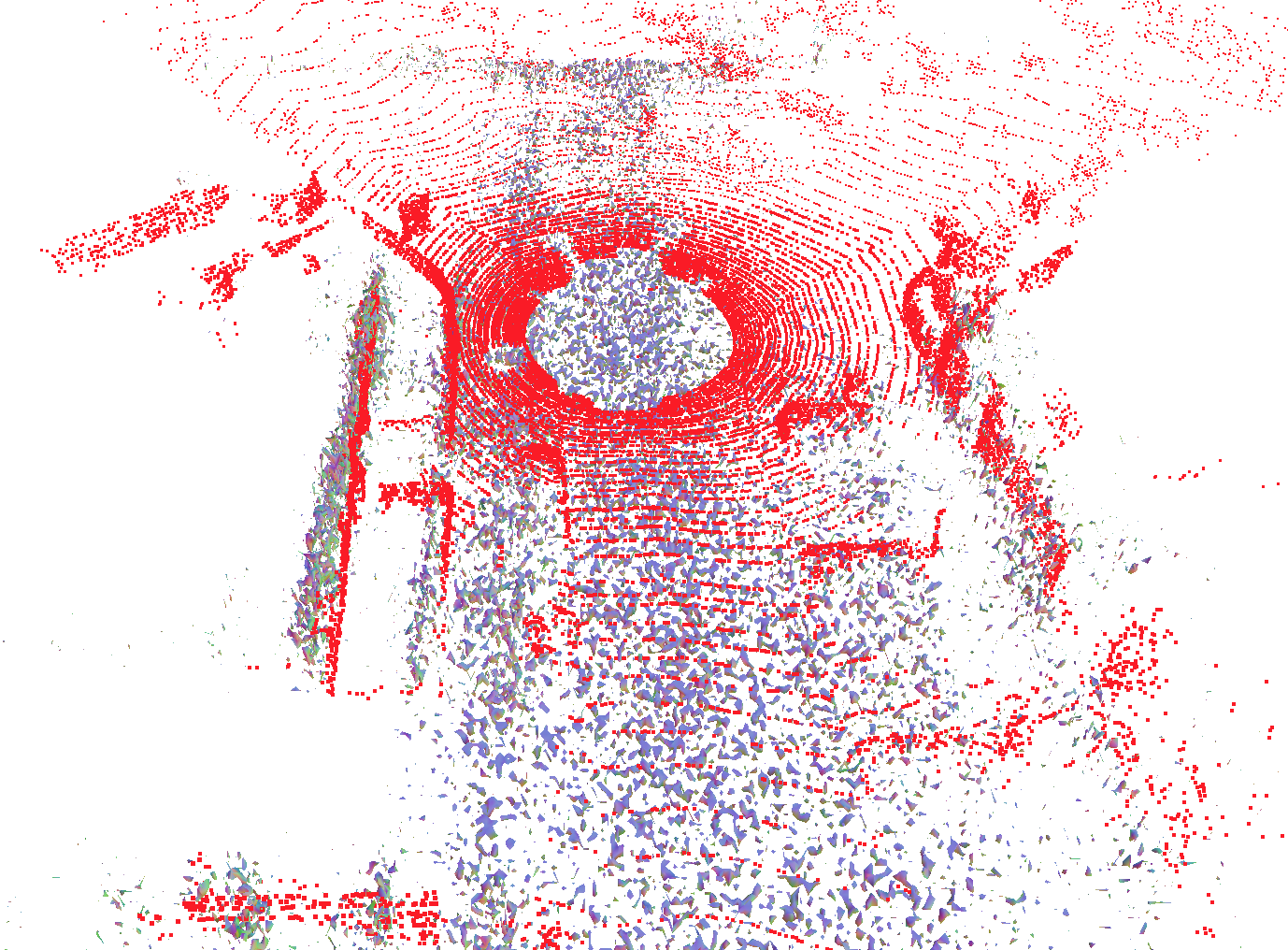}}{SHINE-Mapping}
	\stackunder{\includegraphics[width=0.328\linewidth]{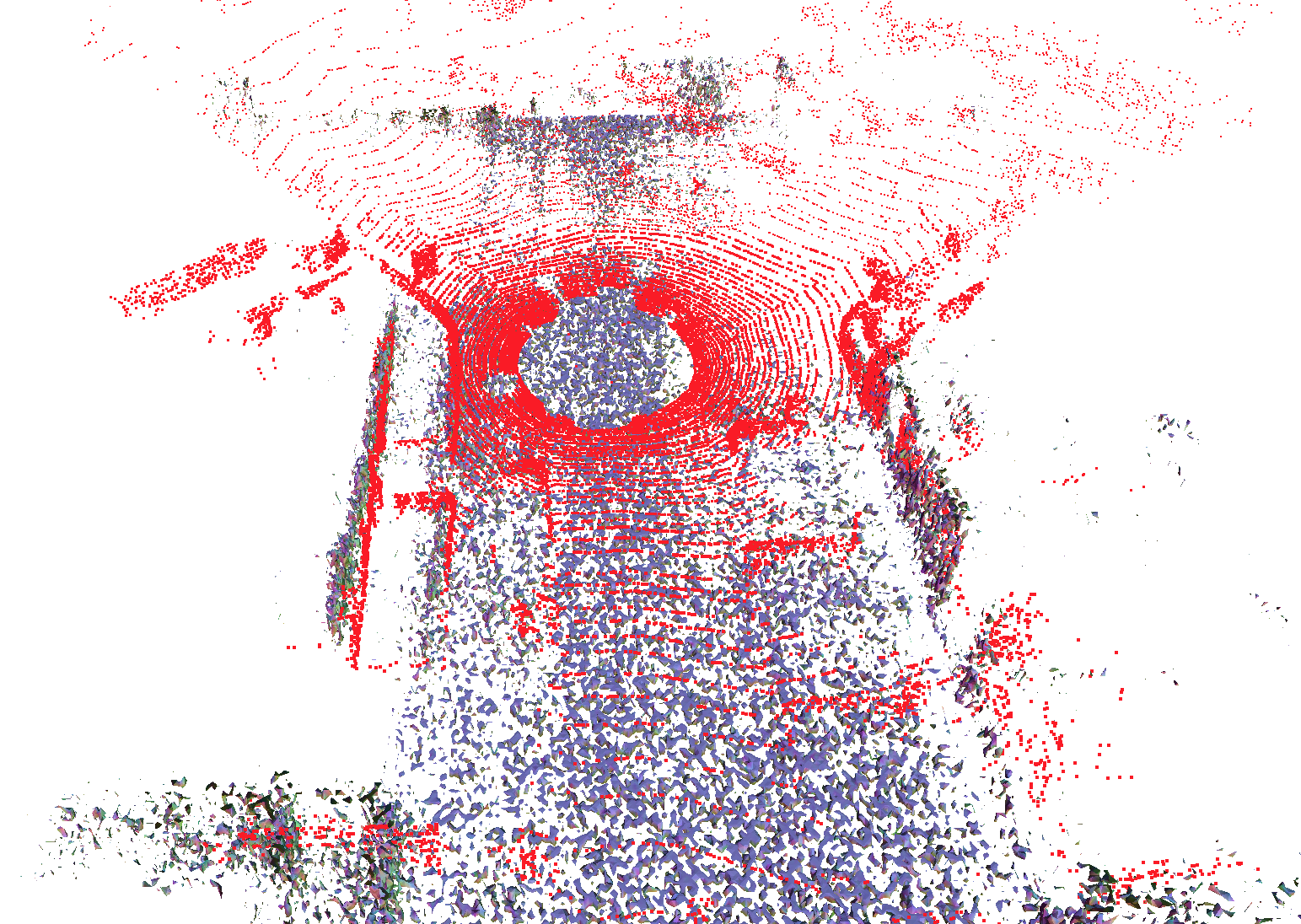}}{LocNDF}
	\stackunder{\includegraphics[width=0.328\linewidth]{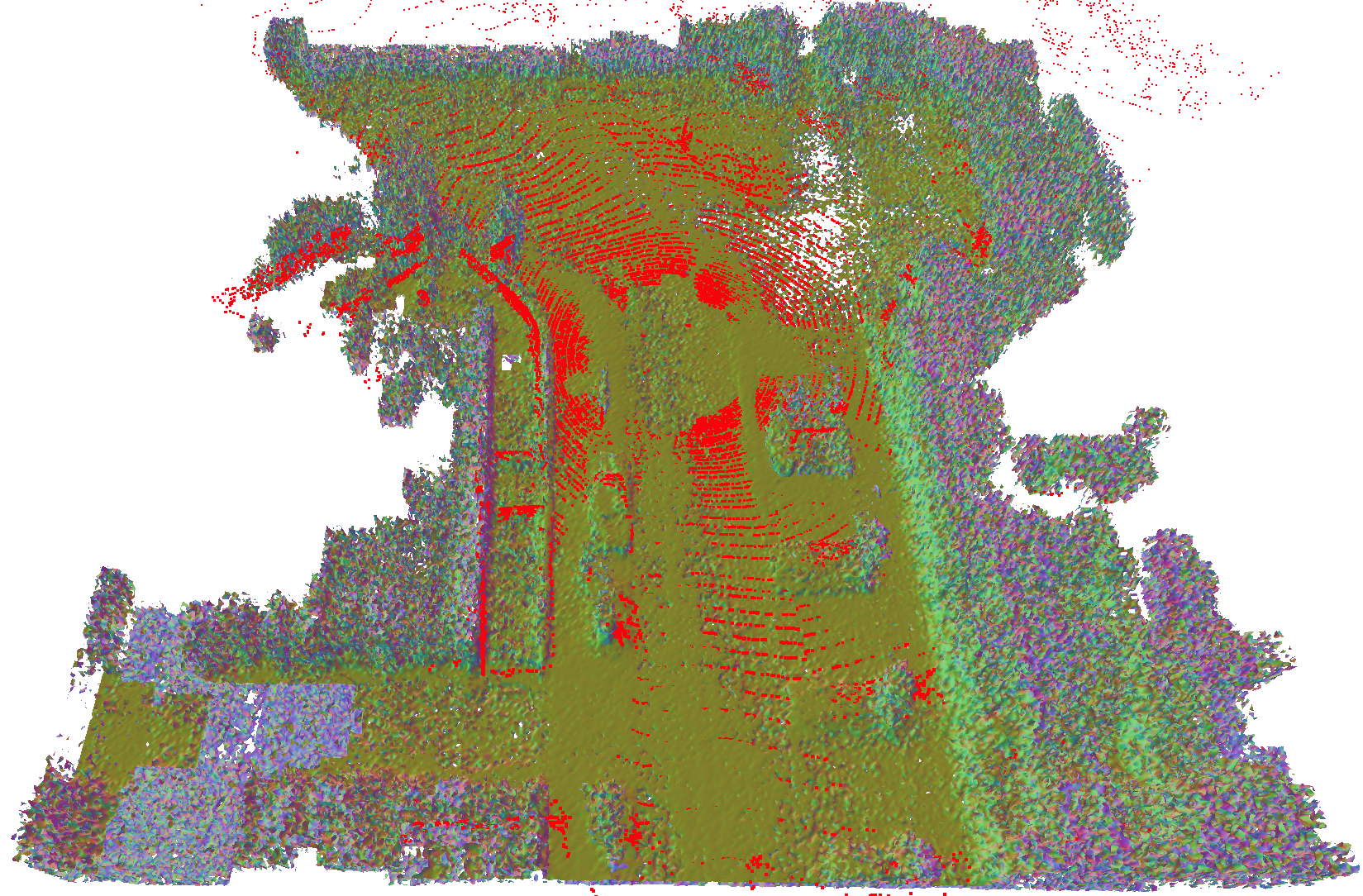}}{CCNDF}
	\caption{Results on KITTI dataset.}
	\label{fkitti}
\end{figure}\begin{figure}
        \stackunder{\includegraphics[width=0.328\linewidth]{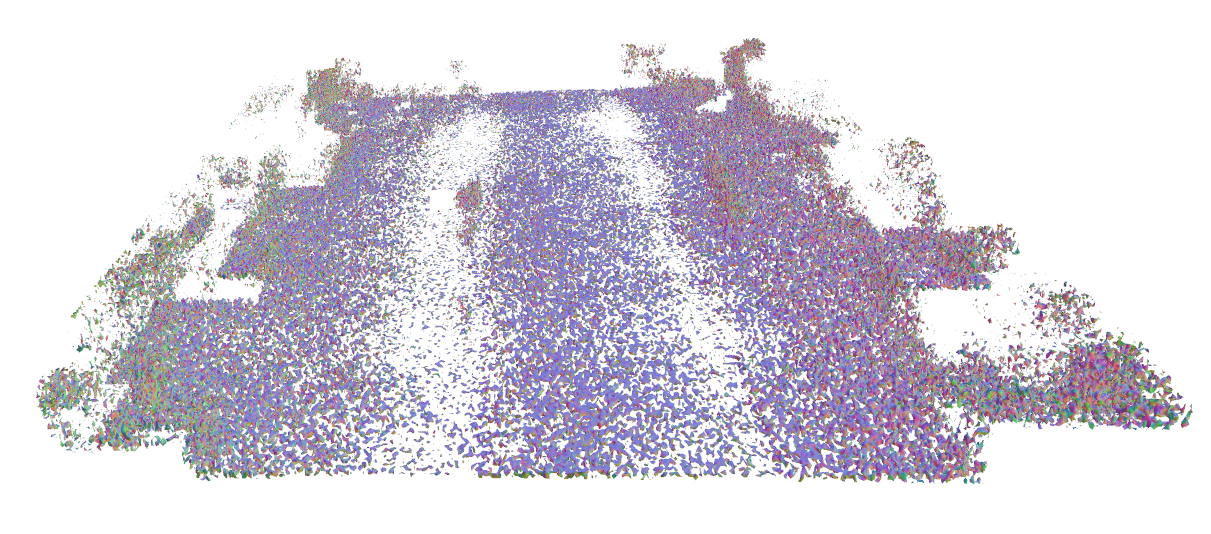}}{(a)}
        \stackunder{\includegraphics[width=0.328\linewidth]{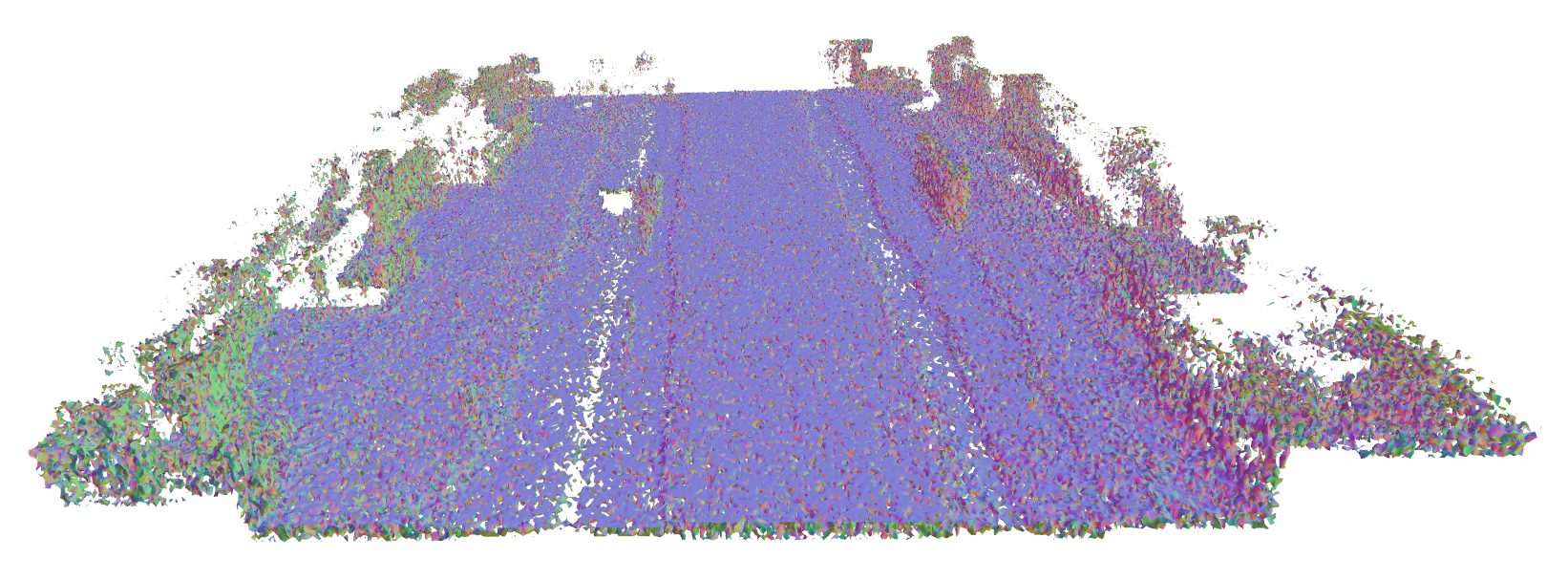}}{(b)}
        \stackunder{\includegraphics[width=0.328\linewidth]{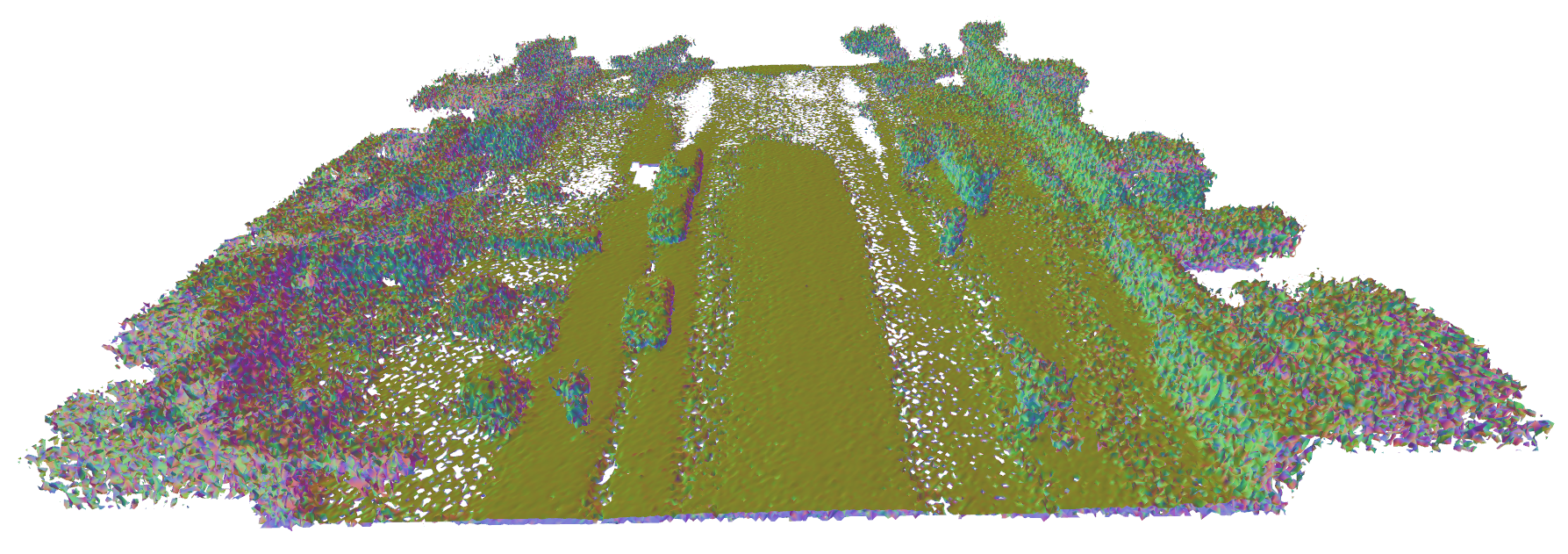}}{(c)}
    \caption{This figure illustrates front-view mappings derived from various geometric methods, with darker colours indicating improved normal estimation. (a) Ray Distance, (b) Ray distance along the normal pointing towards the closest surface, and  (c) Proposed approach.}
    \label{5}
\end{figure} \\
\textbf{Localization}: The rationale behind evaluating NDF based on localization stems from the fact that, while mapping lacks ground truth for evaluation, localization provides a means of comparison with available ground truth locations. The premise is that if localization performs well, it implies good mapping and, by extension, well-constructed NDF.  To quantitatively analyze our methods, we employed global 2D Monte Carlo Localization (MCL). In order to compare our approach, we utilized the sequence as given by Kuang \etal \cite{kuang2023ir}. The particle filter transitions to pose tracking mode with 10,000 particles when the standard deviation of particles falls below 30 cm. The particles are reweighted and resampled if the agent moves by 5 cm or 0.1 radians. The detailed implementation of localization followed the same procedure as employed by the authors of LocNDF \cite{locndf}, ensuring consistency in process and technicality. For method comparison, we utilized root mean squared error (RMSE) and mean average error (MAE) between ground truth and estimated positions. The reported metrics represent averages over 5 runs, with metrics reported only if at least one run converges; otherwise, they are represented as "-". 
We conducted a comparative analysis of our approach against the current SOTA geometrical approach under unified conditions. Table \ref{tab1} presents the comparison between the MCL results obtained by supervising the NDF with expected distances calculated through our proposed method and when using ray distance along the normal pointing towards the closest surface. The results clearly indicate that our approach outperformed the SOTA geometrical approach, as evidenced by lower values of MAE and RMSE. Furthermore, the smaller differences in MAE and RMSE values suggest that our approach generated fewer outliers. The MCL results when the NDF is supervised with ray distance showcase randomised and incoherent outcomes. In many cases, the results fail to converge to pose tracking mode at all. This outcome was anticipated due to the inherent challenges in mapping when supervised by ray distance as seen in the previous subsection. The ineffective mapping adversely impacts localization.\begin{table}[t]
	\caption{MCL Results and Comparative Analysis of Different Geometrical Approaches}
	\label{tab1}
	\centering
	\begin{tabular}{c|c|c|c|c|c|c}
		\hline
		Error Metric & Distance & Seq-1 & Seq-2 & Seq-3 & Seq-4 & Seq-5\\\hline
		& Closest point along normal  & 4.90 & 4.20 & 5.30 & 6.90 & 7.10\\\cline{2-7}
		RMSE & Ray Distance & {\color{blue!85!black}4.42} & {\color{blue!85!black}3.80} & {\color{blue!85!black}4.42} & {\color{green!65!black}\textbf{4.55}} & {\color{green!65!black}\textbf{4.63}}\\\cline{2-7}
		& Proposed Distance & {\color{green!65!black}\textbf{1.40}} & {\color{green!65!black}\textbf{1.80}} & {\color{green!65!black}\textbf{3.40}} & {\color{blue!85!black}5.30} & {\color{blue!85!black}4.90}\\    
		\hline
		& Closest point along normal  & 4.4 & 3.8 & 4.6 & 5.7 & 6.5\\\cline{2-7}
		MAE& Ray Distance & {\color{blue!85!black}1.31} & {\color{blue!85!black}1.97} & {\color{blue!85!black}2.93} & {\color{green!65!black}\textbf{4.57}} & {\color{green!65!black}\textbf{4.57}}\\\cline{2-7}
		& Proposed Distance & {\color{green!65!black}\textbf{1.3}} & {\color{green!65!black}\textbf{1.9}} & {\color{green!65!black}\textbf{2.9}} & {\color{blue!85!black}4.6} & {\color{blue!85!black}4.6}\\
		\hline
	\end{tabular}
\end{table}
To comprehensively evaluate our approach, we conducted tests against various baseline localization methods, including AMCL\cite{fox2001kld}, SRRG \cite{srrg}, IR-MCL\cite{kuang2023ir}, and LocNDF \cite{locndf}. This comparative analysis aimed to provide a more holistic view of our model in comparison to established baselines. The results are reported in Table \ref{tab2}, and our findings suggest that our method outperformed the baseline models. A noteworthy observation from our results is that, unlike other models where RMSE values continue to increase with an increment in the threshold from 5 to 20 cm, our values exhibit convergence and demonstrate a saturating behaviour. We attribute this phenomenon to a better continuous map representation, facilitated by our avoidance of grid resolution limitations and supervision with improved geometrical distances.
\section{Discussion, Limitations and Future work}
\begin{table}[!t]
	\caption{MCL results and comparison of the perfomance of the proposed approach with that of  AMCL \cite{fox2001kld}, SRRG \cite{srrg}, IR-MCL \cite{kuang2023ir}, and LocNDF \cite{locndf}.}
	\label{tab2}
	\centering
	\begin{tabular}{c|c|c|c|c|c|c}
		\hline
		Error Metric & Method & Seq-1 & Seq-2 & Seq-3 & Seq-4 & Seq-5\\
		\hline
		&  AMCL \cite{fox2001kld} & $-$ & 3.7 & 3.8 & 3.4 & $-$\\\cline{2-7}
		& SRRG \cite{srrg} & 3.4 & 3.4 & 3.3 & 3.5 & 3.5\\\cline{2-7}
		RMSE (5cm)& IR-MCL \cite{kuang2023ir} & 3.3 & 2.9 & 3.3 & 3.3 & 3.2\\\cline{2-7}
		& LocNDF \cite{locndf}& {\color{blue!85!black}3.1} & {\color{blue!85!black}2.8} & {\color{green!65!black}\textbf{2.8}} & {\color{green!65!black}\textbf{3.1}} & {\color{green!65!black}\textbf{2.7}}\\\cline{2-7}
		& Proposed & {\color{green!65!black}\textbf{1.4}} & {\color{green!65!black}\textbf{1.9}} & {\color{blue!85!black}2.9} & {\color{blue!85!black}3.4} & {\color{blue!85!black}3.4}\\
		\hline
		&  AMCL \cite{fox2001kld} & $-$ & 6.1 & 6.6 & 4.8 & $-$\\\cline{2-7}
		& SRRG \cite{srrg} & 4.7 & 5.9 & 5.0 & 5.2 & 5.1\\\cline{2-7}
		RMSE (10cm) & IR-MCL \cite{kuang2023ir}& 4.7 & 4.8 & 4.3 & 5.4 & 5.7\\\cline{2-7}
		& LocNDF \cite{locndf}& {\color{blue!85!black}4.1} & {\color{blue!85!black}3.2} & {\color{blue!85!black}3.4} & {\color{blue!85!black}4.6} & {\color{green!65!black}\textbf{3.2}}\\\cline{2-7}
		& Proposed & {\color{green!65!black}\textbf{1.4}} & {\color{green!65!black}\textbf{1.9}} & {\color{green!65!black}\textbf{3.4}} & {\color{green!65!black}\textbf{4.5}} & {\color{blue!85!black}4.9}\\
		\hline
		&  AMCL \cite{fox2001kld}& $-$ & 8.9 & 9.9 & 5.9 & $-$\\\cline{2-7}
		& SRRG \cite{srrg}& 4.9 & 6.3 & 7.5 & 5.8 & 5.5\\\cline{2-7}
		RMSE (20cm) & IR-MCL \cite{kuang2023ir} & 5.2 & 5.4 & 5.4 & 9.1 & 6.4\\\cline{2-7}
		& LocNDF \cite{locndf}& {\color{blue!85!black}4.7} & {\color{blue!85!black}4.2} & {\color{blue!85!black}5.3} & {\color{blue!85!black}6.9} & {\color{blue!85!black}7.1}\\\cline{2-7}
		& Proposed & {\color{green!65!black}\textbf{1.4}} & {\color{green!65!black}\textbf{1.9}} & {\color{green!65!black}\textbf{3.4}} & {\color{green!65!black}\textbf{5.3}} & {\color{green!65!black}\textbf{4.9}}\\
		\hline
	\end{tabular}
	
\end{table}
\textbf{Geometrical Insights Leading to Improved Performance.} Our method excelled due to the effect of the incorporation of expanded geometry, as mentioned. Moreover, the observed superior performance can be attributed to a macroscopic event, which is especially advantageous in large-scale environments. This macro-event is a consequence of the formation and behavior of neural fields over substantial distances. Remarkably, as the distance increases, the neural fields exhibit a tendency to approximate a circular form. This phenomenon is attributed to the corners of the object acting as centres for arcs within the neural field. The intensity of this event increases when the corners are rounded, making this phenomenon more observable in real-world situations. Furthermore, when the inquiry point is positioned at a sufficiently large distance from the surface, the significance of the gradient diminishes while the importance of the ROC remains, providing an accurate scalar value. Notably, for any point situated at a far distance from the surface, the gradient effectively aligns along the ray direction due to the diminishing difference between the ray's endpoint and the closest point on the surface. In contrast, the significance of the ROC persists, remaining dependent on the closest point. This characteristic ensures that the ROC provides an accurate scalar value, a crucial attribute in our context. Notably, with an increase in distance, neural fields tend to adopt a rounded configuration. This effect becomes more prominent in real-world scenarios, especially when the corners are rounded. Additionally, positioning the inquiry point at a considerable distance from the surface reveals an interesting dynamic. Here, the significance of the gradient diminishes while the importance of the ROC persists, providing a reliable scalar value. Specifically, for points at a significant distance from the surface, the gradient effectively aligns along the ray direction due to the diminishing difference between the ray's endpoint and the closest point on the surface. In contrast, the significance of the ROC remains, maintaining dependence on the closest point. This characteristic ensures that the ROC furnishes an accurate scalar value, a crucial attribute within our context.\\
Despite employing a logarithmic linear sampling strategy that assigns heightened importance to distances computed near the surface, the observed significance of distances calculated from distant points remains crucial. This observed phenomenon not only contributes to the stability of our training process but also strengthens the credibility of our approach. Additionally, it underscores the pivotal role of the ROC in effectively supervising the NDF. \\
\textbf{Geometric Limitation.} The sole geometric constraint encountered in our methodology arises when the LiDAR ray fails to intersect the ROC at the closest point on the surface. For instance, in Fig. \ref{fig2:overall}(a), if the ray from $A$ at $\mathsf{e}_i$ does not intersect the ROC at point $F$, the calculated ROC at $F$ may introduce errors. However, the practical significance of this limitation is mitigated by our approach's emphasis on sampling more points in close proximity to the surface. In large-scale environments, where the ROC of objects is sufficiently large, incoming rays intersect the ROC at the closest point, minimizing the impact of this limitation. Additionally, for points farther from the surface, the NDF behaves akin to a sphere, further mitigating the consequences of this limitation. The alignment of this limitation with its predecessors is inherent, as the sampling of points near the surface is a common practice in effectively supervising the NDF across various methodologies.\\
\textbf{Future Work.} This work was conducted within an offline setting, and the prospect of extending this approach to an online setting presents an interesting future aspect. Additionally, the exploration of alternative deep learning models to model the NDF holds promise for further advancing the understanding and capabilities.

\section{Conclusion} 
In this work, we introduced a novel geometric approach for supervising NDF. Our method was evaluated on two pivotal problems—mapping and localization, where NDF plays a fundamental role. We compared our approach with state-of-the-art geometric methods commonly used for NDF supervision. The results indicate that our method surpasses current techniques in performance and is more geometrically and mathematically aligned with the inherent properties of NDF. By leveraging higher-order properties of NDF for supervision, we established a foundation for further exploration of these properties to enhance NDF's versatility in various tasks. Our methodology advances the field and opens avenues for future research into higher-order properties of NDF.
\bibliographystyle{splncs04}
\bibliography{main}
\end{document}